\documentclass[10pt,twocolumn,letterpaper]{article}

\usepackage{cvpr}
\usepackage{times}
\usepackage{epsfig}
\usepackage{graphicx}
\usepackage{amsmath}
\usepackage{amssymb}
\usepackage{blindtext}
\graphicspath{{./}}
\usepackage{ctable}
\DeclareGraphicsExtensions{.pdf,.jpeg,.png,.eps,.jpg,.jpeg}
\usepackage{multirow}
\usepackage{hhline,xcolor}
\usepackage{makecell}
\usepackage{boldline}

\usepackage[pagebackref=true,breaklinks=true,letterpaper=true,colorlinks,bookmarks=false]{hyperref}

\usepackage[pagebackref=true,breaklinks=true,letterpaper=true,colorlinks,bookmarks=false]{hyperref}

\cvprfinalcopy 

\def\cvprPaperID{5669} 

\ifcvprfinal\fi
\begin{document}
	
	\title{Joint Spatial-Temporal Optimization for Stereo 3D Object Tracking}
	
\author{Peiliang Li, Jieqi Shi, and Shaojie Shen\\
		The Hong Kong University of Science and Technology\\
{\tt\small pliap@connect.ust.hk, jshias@connect.ust.hk, eeshaojie@ust.hk}
}		
	\maketitle
	\thispagestyle{empty}
	
	\begin{abstract}
		Directly learning multiple 3D objects motion from sequential images is difficult, while the geometric bundle adjustment lacks the ability to localize the invisible object centroid.
		To benefit from both the powerful object understanding skill from deep neural network meanwhile tackle precise geometry modeling for consistent trajectory estimation, we propose a joint spatial-temporal optimization-based stereo 3D object tracking method.
		From the network, we detect corresponding 2D bounding boxes on adjacent images and regress an initial 3D bounding box. Dense object cues (local depth and local coordinates) that associating to the object centroid are then predicted using a region-based network.
		Considering both the instant localization accuracy and motion consistency, our optimization models the relations between the object centroid and observed cues into a joint spatial-temporal error function. All historic cues will be summarized to contribute to the current estimation by a per-frame marginalization strategy without repeated computation.
		Quantitative evaluation on the KITTI tracking dataset shows our approach outperforms previous image-based 3D tracking methods by significant margins. We also report extensive results on multiple categories and larger datasets (KITTI raw and Argoverse Tracking) for future benchmarking.
	\end{abstract}
	
	\section{Introduction}
	3D object detection and tracking play a significant role for autonomous driving vehicles where the time-independent detection undertakes the fundamental perception, and continuous object tracking further enables temporal motion prediction and planning. With the rapid evolution of 3D deep learning and feature representation, the detection part has been made great progress in terms of 3D localization ability by many efforts \cite{cvpr18xu, ku2019monocular, ma2019accurate, li2019stereo, wang2019pseudo, chen2017multi,qi2017frustum,zhou2017voxelnet,liang2018deep,liang2019multi, lang2019pointpillars, shi2019pointrcnn}. However, as an equally essential task, the 3D object tracking has rarely been explored. Only a few recent works \cite{li2018stereo, luo2018fast, hu2018joint} demonstrate 3D object tracking ability in the context of self-driving scenarios. To bridge this gap and take advantage of sequential visual cues, we aim at a complete 3D object tracking system that joint exploits spatial-temporal information and estimates accurate 3D object trajectories with motion consistency. We focus on the use of stereo cameras as it shows a promising balance between the cost and 3D sensing ability comparing with the expensive LiDAR sensor and the inadequate single camera.
	
	\begin{figure}
		\setlength{\belowcaptionskip}{-0.8cm} 
		\begin{center}
			\includegraphics[width=1.0\columnwidth]{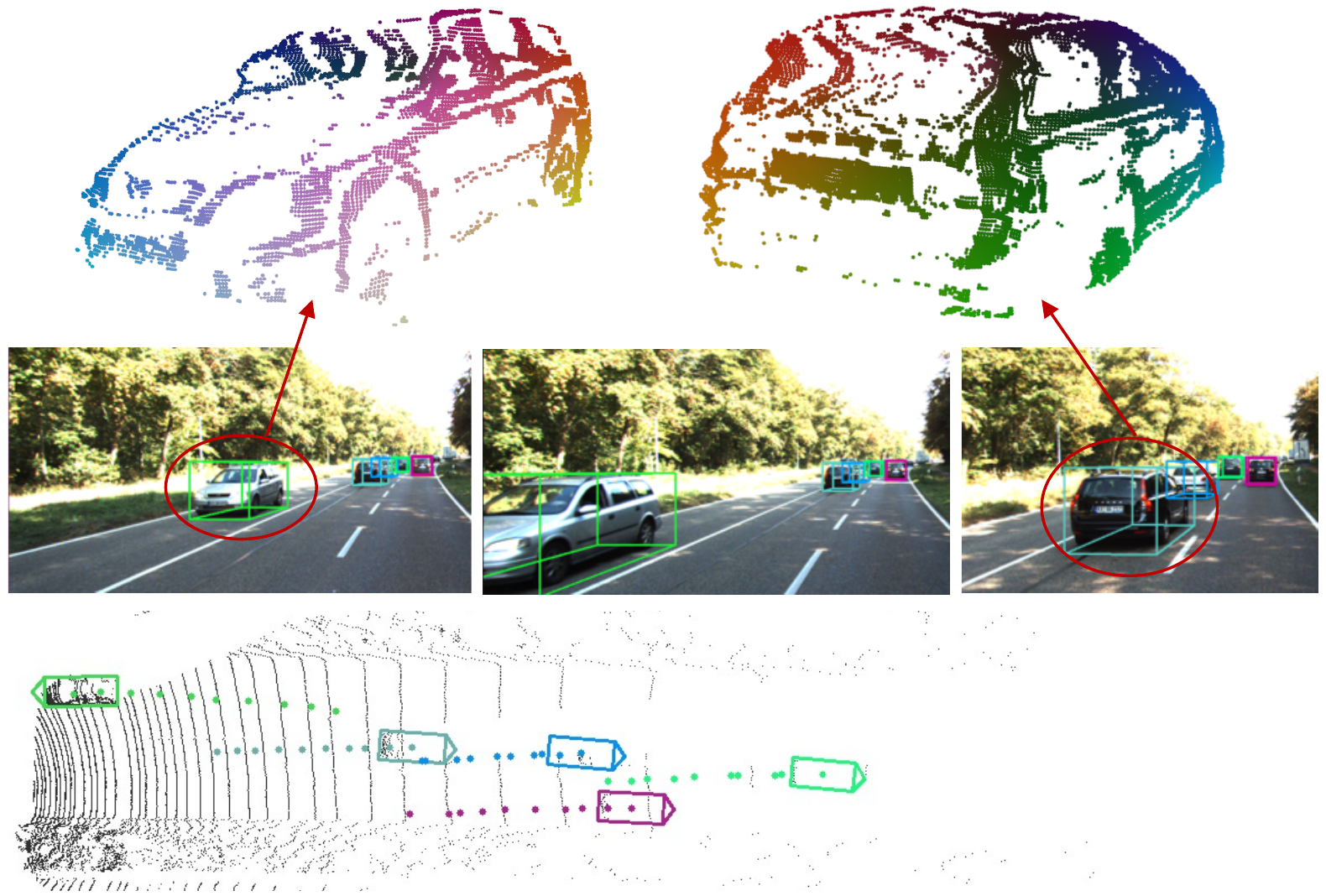}
		\end{center}
		\vspace{-0.5em}
		\caption{\textbf{An example of our 3D tracking system.} From top to bottom: The sampled object local depth which is color mapped by local coordinates; 3D tracking result on sequential images; 3D tracking result on the bird's eye view. Here the trajectory is transformed to global coordinates for visualization using the off-shelf ego-camera pose.}
		\vspace{-0.5em}
		\label{fig:cover}
	\end{figure}

	\begin{figure*}
		\begin{center}
			\includegraphics[width=1.91\columnwidth]{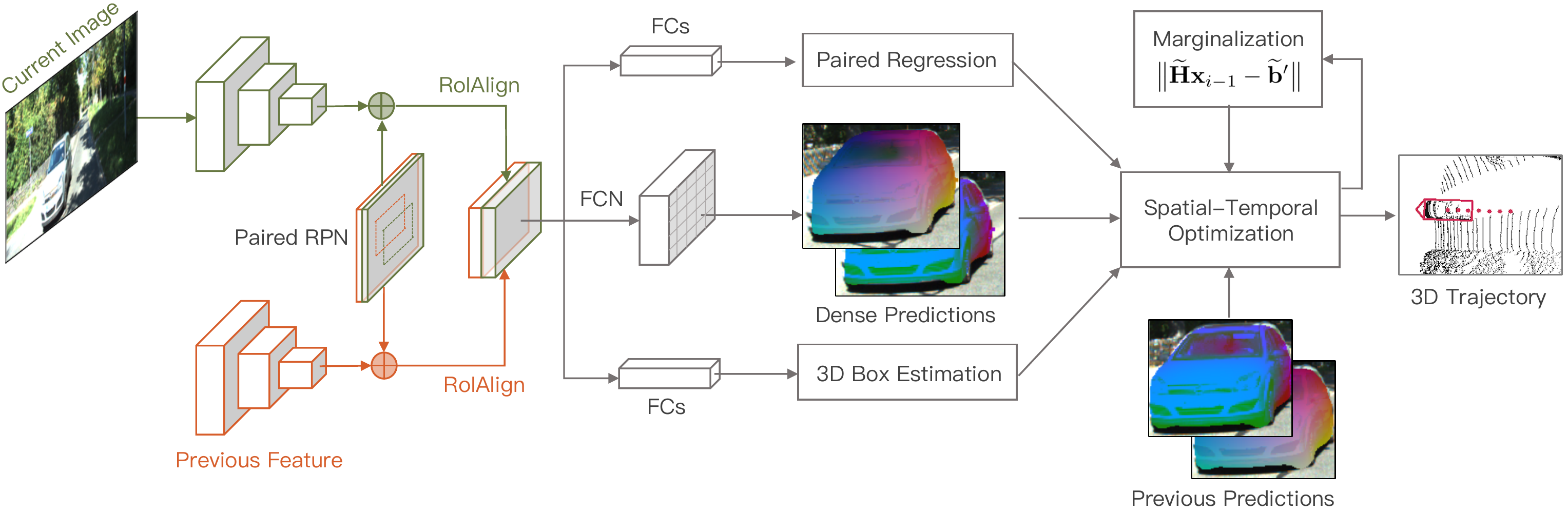}
		\end{center}
		\caption{Architecture of the proposed Stereo 3D object tracking system, which generates paired 2D boxes for data association (Sect.~\ref{sec:det}), initial 3D box estimation (Sect.~\ref{sec:box}) and dense local predictions (Sect.~\ref{sec:dense}) for the following spatial-temporal optimization (Sect.~\ref{sec:opti}).}
		\label{fig:system}
	\end{figure*}
		
	In this paper, we consider the 3D object tracking as not only a \textit{data association} problem but also a \textit{continuous state estimation} problem, where the estimation result should satisfy both instant spatial constraints and the accumulated history evidence. To this end, we propose our joint spatial-temporal optimization-based 3D object tracking system. As illustrated in Fig.~\ref{fig:system}, 
	our system firstly generates paired region proposals on the concatenated current and previous feature maps. After \textit{RoIAlign} \cite{he2017mask}, we employ three parallel branches on the concatenated RoI (region of interest) feature to refine the proposal and generate object-level and pixel-level information. As Fig.~\ref{fig:system} shows, the paired regression branch refines the paired proposals to accurate 2D bounding box pairs. Benefit from the paired detection, the sequential objects can be naturally associated without additional similarity computation (Sect.~\ref{sec:det}). 
	The 3D estimation branch predicts object-level information, e.g., centroid projection, observation angle to form an initial 3D box (Sect.~\ref{sec:box}). The dense prediction branch outputs pixel-level segmentation and local geometry cues such as local depth and local coordinates (Sect.~\ref{sec:dense}) that are aggregated later to formulate our spatial-temporal optimization.
		
	To estimate a consistent and accurate 3D trajectory, we enforce precise geometry modeling by jointly considering the dense spatial and historic cues. From the spatial view, an optimal object depth should yield minimal stereo photometric error given the local depth relations between foreground pixels and object centroid. From the temporal view, the consistent object motion will yield minimal reprojection error after warping foreground pixels to the adjacent frame. Based on this, we propose a joint spatial-temporal optimization which models all these dense constraints in a tightly-coupled manner (Sect.~\ref{sec:opti}). To trade off the large amount information introduced by dense cues from multiple frames, we further introduce a per-frame marginalization strategy where the previous observations will be iteratively marginalized as a linear prior, s.t., all historical evidence will naturally contribute to the current object estimation without the need of information reuse.

	Overall, our main contributions can be summarized as:
	\begin{itemize}
		\item A complete 3D object tracking framework that handles simultaneous detection \& association via learned correspondences, and solves continuous estimation by fully exploiting dense spatial-temporal constraints.
		\item Significantly outperform state-of-the-art image-based 3D tracking methods on the KITTI tracking dataset.
		\item Report extensive evaluation on more categories and larger-scale datasets (KITTI Raw and Argoverse Tracking) to benefit future benchmarking.
	\end{itemize}

	\section{Related Work}
	{\setlength{\parindent}{0cm}
		\subparagraph*{3D Object Detection.} There are plenty of research efforts focus on the detecting 3D object using instant sensor data in autonomous driving scenarios. From the modality of input data, we can roughly outline them into three categories: monocular image-based methods,
		stereo imagery-based,
		and LiDAR-based methods.
		Given a \textbf{monocular image}, some earlier works \cite{chen2016monocular, zeeshan2014cars, chabot2017deep, kundu20183d} exploit multiple levels of information such as segmentation, shape prior, keypoint, and instance model to help the 3D object understanding, while recent state-of-the-art works \cite{cvpr18xu, qin2019monogrnet, ku2019monocular, ma2019accurate, brazil2019m3d} pay more attention to the depth information encoding from different aspects to detect and localize the 3D object. Adding additional images with known extrinsic configuration, \textbf{stereo based methods} \cite{3dopJournal, li2019stereo, wang2019pseudo, pon2019object} demonstrate much better 3D object localization accuracy, where \cite{li2019stereo} utilizes object-level prior and geometric constraints to solves the object pose using raw stereo image alignment. \cite{wang2019pseudo} converts the stereo-generate depth to a pseudo point cloud representation and directly detect object in 3D space. While \cite{pon2019object} takes advantages of both and predict object-level point cloud then regress the 3D bounding box based on the object point cloud. Besides the image-based approaches, rich works \cite{li2016vehicle, engelcke2017vote3deep, zhou2017voxelnet, lang2019pointpillars, yang2018pixor, qi2017frustum, shi2019pointrcnn} utilize the direct 3D information from the \textbf{LiDAR point cloud} to detect 3D objects, where \cite{engelcke2017vote3deep, zhou2017voxelnet, lang2019pointpillars} samples the unstructured point cloud into structured voxel representation and use 2D or 3D convolution to encode features. \cite{li2016vehicle, yang2018pixor} project the point cloud to the front or bird's eye views such that the 3D object detection can be achieved by regular 2D convolutional networks. From another aspect, \cite{qi2017frustum, shi2019pointrcnn} directly localize 3D objects in unstructured point cloud with the aid of PointNet \cite{qi2017pointnet} encoder. Furthermore, \cite{chen2017multi, ku2017joint, liang2018deep, liang2019multi} exploit fuse the image and point cloud in feature level to enable multi-modality detection and scene understanding.
	}
	\vspace{-0.2cm}
	{\setlength{\parindent}{0cm}
		\subparagraph*{3D Object Tracking.} Although extensive object tracking approaches have been studied in recent decades, in this paper, we mainly discuss the most relevant literature: the 3D object tracking. Based on the 3D detection results, \cite{osep2017combined, scheidegger2018mono, simon2019complexer} employ a filter based modeling (multi-Bernoulli, Kalman filter) to track the 3D object continuously. Alternatively, \cite{hu2018joint} directly learns the object motion using an LSTM model by taking advantage of data-driving approaches. However, decoupling the detection and tracking might cause a sub-optimal solution due to the information loss. Benefit from the stereo vision, \cite{engelmann2017samp} focuses on the object reconstruction with continuous tracked visual cues, and \cite{li2018stereo} employ an object bundle adjustment approach to solve consistent object motion in a sliding window, while relying on the sparse feature matching and loosely coupling the stereo reconstruction with temporal motion limits its performance on 3D localization for occluded and faraway objects. In another way, \cite{luo2018fast} encodes sequential 3D point cloud into a concatenated voxel representation, and directly produces associated multi-frame 3D object detections as tracked trajectory together with motion forecasting.
	}
	
	\section{Sequential 3D Tracking Network}
	In this section, we describe our sequential 3D object tracking network, which simultaneously detects and associates objects in consecutive monocular images, and introduce our network predictions to enable the initial 3D box estimation and the subsequent joint optimization (Sect.~\ref{sec:opti}). 
	
	\subsection{Simultaneous Detection \& Association}
	\label{sec:det}
	To avoid additional pair-wise similarity computation for object association, we leverage the network directly detect the corresponding objects in adjacent frames. Before object-wise estimation, we use the region proposal network (RPN) \cite{ren2015faster} to densely classify the foreground object and predict coarse object region on the feature map. Inspired from \cite{li2019stereo} for stereo detection, we extend the RPN to recognize the union area of where the object appearing in two sequential images. Specifically, 
	After feature extraction, the feature maps of current image and the previous image are concatenated to involve temporal information (see Fig.~\ref{fig:system}).
	We pre-calculate the union of corresponding object boxes in the current and previous images. 
	On the concatenated feature map with per-location defined anchors, an anchor will be classified as the foreground only if its IoU (intersection-over-union) with one of the union box is larger than 0.7. 
	On this definition, the positive anchor will cover the object area on both images, thereby it can be regressed to paired RoIs proposals on the current and previous image respectively. 
	Note that this paired RPN does not beyond the network capability since it can be thought as a special classification task where only a repeated and motion-reasonable pattern with same instances can be recognized as a positive sample.
	
\begin{figure}
		\begin{center}
			\includegraphics[width=1.0\columnwidth]{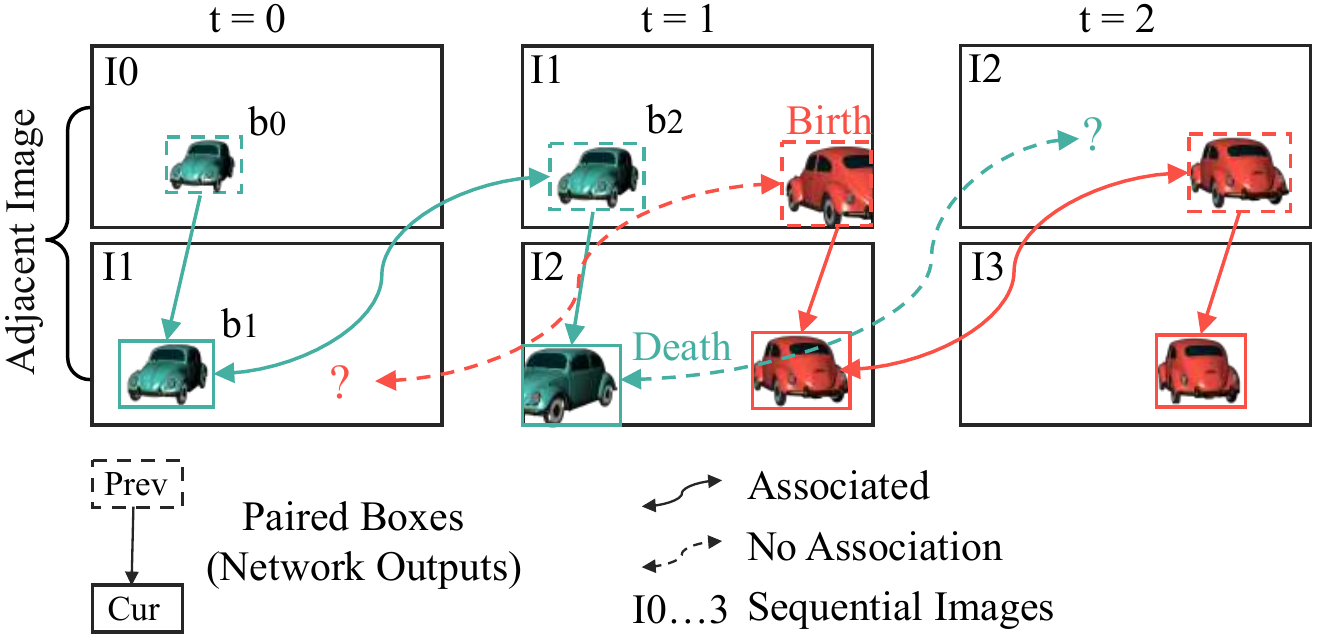}
		\end{center}
		\caption{\textbf{Association illustration}. At t0 timestamp, the network predicts paired 2D box (b0, b1) for image I0 and I1 respectively. Then at time T1, the green car can be associated by comparing 2D IoU of b1 and b2. The \textit{Birth} and \textit{Death} represent the newborn trackers and died trackers respectively.}
		\label{fig:ass}
\end{figure}
	The coarse proposal pairs are further refined in the paired regression.  As Fig.~\ref{fig:system} shows, we use the RoI pairs to perform \textit{RoIAlign} \cite{he2017mask} on the current and previous feature maps respectively. The cropped current and previous RoI features are then concatenated again to enable the R-CNN based 2D box refinement. By predicting two sets of box offset $[\Delta x, \Delta y, \Delta w, \Delta h]$ which denote offsets in $x, y$ direction, width and height, we obtain paired 2D boxes for current and previous images at each timestamp. During inference, we associate sequential trackers by comparing the 2D IoU between the previous box and current's previous box. An example is visualized in Fig.~\ref{fig:ass} for better illustration.
	
	Benefited from this simple design, we achieve simultaneous object detection and association with almost no additional computation, and avoid being affected by large motion as the neural network can find the correspondences around large receptive field.

	\subsection{3D Object Estimation}
	\label{sec:box}
	A complete 3D bounding box is parameterized by $[x,y,z,w,h,l,\theta]$, where $x,y,z$ are the 3D centroid position respecting to the camera frame, $w,h,l$ the 3D dimension, and $\theta$ the horizontal orientation. Since the global location information is lost after crop and resize operation in \textit{RoIAlign} \cite{he2017mask}, we predict several local geometric properties (centroid projection, depth, observation angle $\alpha$) to form a initial 3D bounding box. The centroid projection is defined as the projection coordinates of the 3D object centroid on the image, which can be learnt from the offset between the projection center and the RoI center. For dimension and depth, we predict a residual term based on a dimension prior and an inferred coarse depth $f\frac{h_{3d}}{h_{2d}}$, given by the focal length $f$, 3D object height $h_{3d}$, and 2D RoI height $h_{2d}$. The observation angle represents the object local viewpoint, which can be learnt from the deformed RoI pattern. Note that the observation angle $\alpha$ is not equivalent to the object orientation $\theta$, instead holds the relation: $\alpha = \theta + \arctan\frac{x}{z}$, as proved in \cite{mousavian20173d, li2019stereo}.

	\subsection{Dense Prediction}
	\label{sec:dense}
	However, the predicted 3D box is far from enough for a consistent and accurate 3D tracker as it does not explicitly utilize spatial nor temporal information, we thus define essential dense representations to enable our following joint spatial-temporal optimization.
	
	{\setlength{\parindent}{0cm} 
		\textbf{Mask:} We use a stacked region-based FCN layers \cite{he2017mask} to predict dense object mask on the RoI feature maps, which is used for our foreground pixel selection.
	}
	
	{\setlength{\parindent}{0cm} 
		\textbf{Local Depth:} For the foreground pixel, we define the local depth $\delta$, given by the depth difference between the pixels and the object centroid, which are integrated later for constraining the object centroid depth using stereo alignment.
	}
	
	{\setlength{\parindent}{0cm} 
		\textbf{Local Coordinates:} We predict each pixel's 3D local coordinates respecting to the object frame as also used in \cite{wang2019normalized, li2019multisensor}. On this representation, the same part of the object holds a unique coordinate which is invariant with object translation and rotation across time, therefore it can be used as the geometric descriptor to obtain dense pixel correspondences between sequential object patches. Comparing to traditional descriptor such as ORB \cite{rublee2011orb}, our learned local coordinate is a unique representation in object domain and robust to perspective-changing, textureless, and illumination variance, thereby give us robust and dense pixel correspondences even for high occluded and faraway objects.
	}
	
	\section{Joint Spatial-Temporal Optimization}
	\label{sec:opti}
\begin{figure}
	\begin{center}
		\includegraphics[width=1.0\columnwidth]{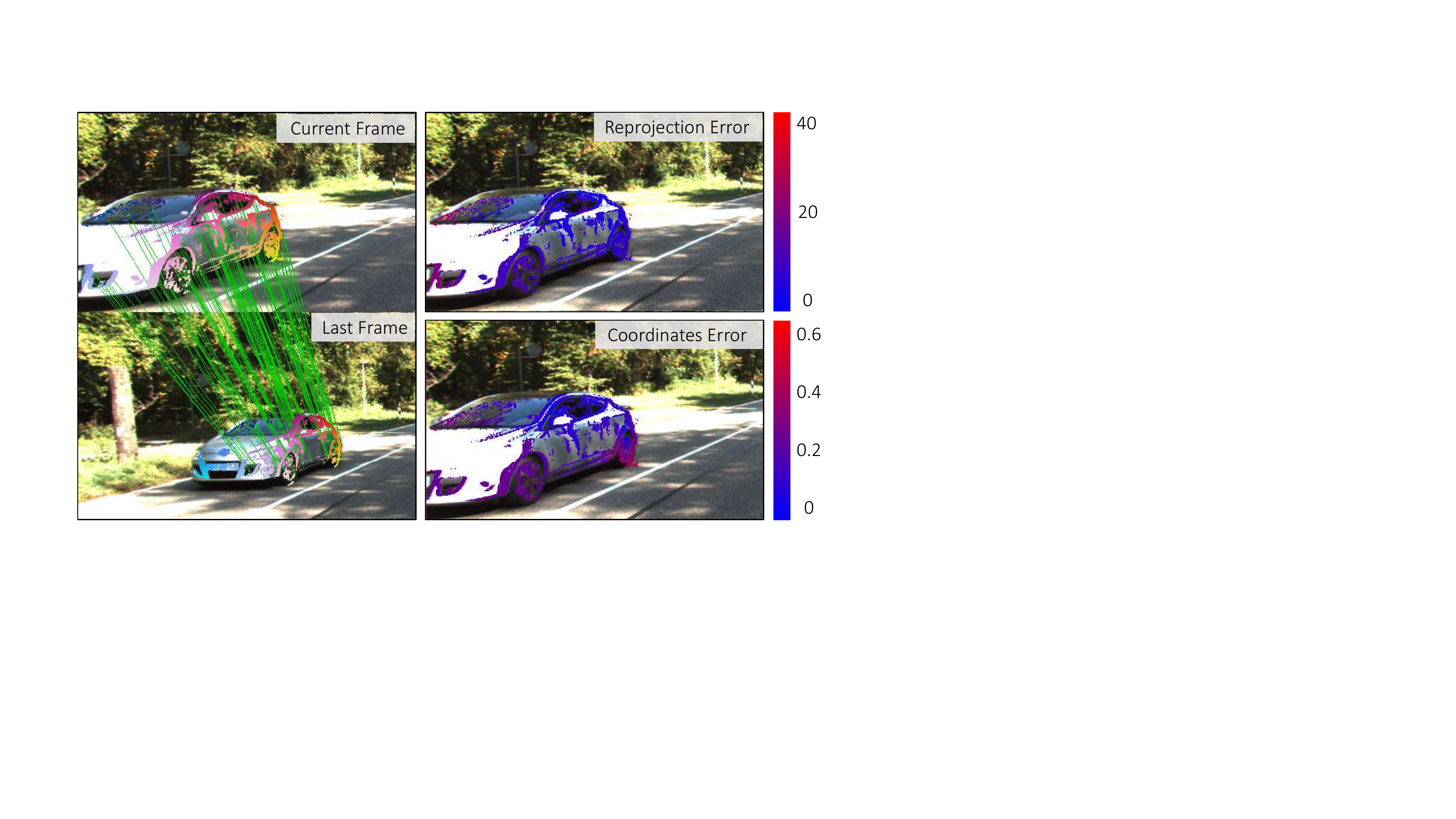}
	\end{center}
	\caption{\textbf{Left}: Pixel correspondences, where we overlay the color mapped local coordinates on adjacent images. Green lines show sampled pixel matches using pair-wise coordinates distance. \textbf{Right}: {Reprojection error vs. Coordinates error}. The top and bottom show the error pattern for reprojection and coordinates aligning respectively.}
	\label{fig:error}
\end{figure}

	Based on these network predictions, we introduce our joint spatial-temporal optimization model. For simplicity, we consider a single object in the following formulation since we solve all objects analogously in a parallel way. Let $I^l_i, I^r_i$ be the sequential stereo image where $i$ denote the frame index, $\mathbf{c}_i$,  $\alpha_i$ be the predicted object centroid projection and observation angle respectively. Let $\mathbf{u}_i$ be the observed foreground pixels given by the object mask. For each observed pixel, we have the local depth $\delta_i$ which serves as spatial cues for stereo alignment, local coordinates $C_i$ that serve our temporal cues for pixel association. For each object, we aim to estimate an accurate object position $\mathbf{p}_i=\{x,y,z\}$ and rotation $\mathbf{R}_i(\theta)$ respecting to the instant camera frame, s.t., we have overall minimum spatial alignment errors and meanwhile are best fitted with the previous pixel observation across multiple frames. 
	\vspace{-0.2cm}
	{\setlength{\parindent}{0cm}
		\subparagraph*{Spatial Alignment.}The spatial alignment cost is defined as the photometric error between left-right images:
		\begin{equation}
		\label{eq:error1}
		\begin{array}{lr}
		\mathbf{E}_{si} := \displaystyle\sum_{\mathbf{u}_i\in\mathcal{N}_s}{w_I} \Big\| I^l_i(\mathbf{u}_i)- I^r_i({\mathbf{u}^r_i}) \Big\|_h,
		\end{array}
		\end{equation}
		where $\mathcal{N}_s$ is the set of sampled foreground pixels according to the image gradient, $w_I$ the weight of the photometric error, $\|\cdot\|_h$ the Huber norm. $\mathbf{u}^i_r$ represents the warped pixel location on the right image $I^r_i$, given by
		\begin{equation}
		\label{eq:error11}
		\begin{array}{lr}
		\mathbf{u}^r_i = \pi \big(\pi^{-1}(\mathbf{u}_i, \delta_i + \mathbf{p}^z_i) + \mathbf{p}_s\big)
		\end{array}
		\end{equation}
		where we use $\pi(\mathbf{p})$ $:$ $\mathbb{R}^3$ $\rightarrow$ $\mathbb{R}^2$ to denote projecting a 3D point $\mathbf{p}$ on the image and $\pi^{-1}(\mathbf{u}, d)$ $:$ $\mathbb{R}^2 \times \mathbb{R}$ $\rightarrow$ $\mathbb{R}^2$ its back-projection according to the pixel localtion $\mathbf{u}$ and depth $d$. The per-pixel depth is given by the predicted $\delta_i$ and the object centroid depth $\mathbf{p}^z_i$, i.e., all pixels are associated with the object centroid. $\mathbf{p}_s$ stands for the extrinsic translation between stereo cameras. Note that we formulate a more accurate stereo alignment model using our predicted local depth (see Fig.~\ref{fig:cover}) instead of the na\"{\i}ve box-shape in \cite{li2019stereo}.
	}
	\vspace{-0.2cm}
	{\setlength{\parindent}{0cm}
		\subparagraph*{Temporal Constraints.}
		Benefit from the geometric and unique property of the local coordinates representation, we can easily obtain temporal pixel correspondences by calculating the pairwise Euclidean distance between the local coordinates in associated object patches. An example of pixel correspondences can be found in the left column of Fig.~\ref{fig:error}. Let $\mathbf{u}_{i-1}$ be the dense correspondences for $\mathbf{u}_{i}$ in the previous frame, given by selecting the closest local coordinates. The temporal constraints encourage all $\mathbf{u}_{i}$ should also be projected to $\mathbf{u}_{i-1}$ (minimal reprojection error) after rigid-body transformation. Let
		\begin{equation}
		\label{eq:error2}
		\begin{array}{lr}
		\mathbf{E}_{ti} := \displaystyle\sum_{\mathbf{u}_i\in\mathcal{N}_t}w_{\mathbf{p}} \Big\| \mathbf{u}^p_{i} - \mathbf{u}_{i-1} \Big\|_h,
		\end{array}
		\end{equation}
		where $\mathcal{N}_t$ is the set of pixels which found correspondence in the previous frame. $\mathbf{u}^p_{i}$ stands for the projected position of $\mathbf{u}_i$ in the previous frame, given by
		\begin{align*}
		\label{eq:error21}
		\begin{array}{lr}
		\mathbf{u}^p_i = \pi \Big(\mathbf{R}_{i-1}\mathbf{R}^{-1}_i\big(\pi^{-1}(\mathbf{u}_i, \delta_i +  \mathbf{p}^z_i) - \mathbf{p}_i\big) + \mathbf{p}_{i-1}\Big).
		\end{array}
		\end{align*}
	}
	\vspace{-0.6cm}
	{\setlength{\parindent}{0cm}
		\subparagraph*{Pose Error.}
		In above equations, the object pose in consecutive frames are coupled together, i.e., only relative motion is constrained. Although the object depth $\mathbf{p}^z_i$ is fully observable from Eq. \ref{eq:error11}, we still need object-level observation angle $\alpha_i$ and centroid projection $\mathbf{c}_i$ to constrain the object orientation and position in $x, y$ direction for each frame separately, which can be simply given by a linear error
		\begin{equation}
		\label{eq:error_pose}
		\begin{array}{lr}
		\mathbf{E}_{pi} := \| \pi(\mathbf{p}_i) - \mathbf{c}_i\| + \| \theta_i - \alpha_i - \arctan(\frac{\mathbf{p}^x_i}{\mathbf{p}^z_i})\|
		\end{array}
		\end{equation}
	}
	{\setlength{\parindent}{0cm}
		\subparagraph*{Per-Frame Marginalization.} To utilize the history information, a straight forward solution would be minimizing all above error terms over multiple frames in a sliding window,
		\begin{equation}
		\label{eq:error_s}
		\begin{array}{lr}
		\mathbf{E}_n = \displaystyle\sum_{i = 0}^n \mathbf{E}_{si} + \mathbf{E}_{ti} + \mathbf{E}_{pi}.
		\end{array}
		\end{equation}
However, re-evaluating the spatial alignment cost for all history frames at each timestamp is unnecessary as we already reach the minimum photometric error for historic frames. To fully exploit history information while avoiding the repeated computation, we use a per-frame marginalization strategy to convert the information from the previous optimization to a prior knowledge for the current frame, which is a common technique in SLAM approaches \cite{engel2017direct, qin2018vins}.}	
	
	For each new object, we joint solve the first two frames by minimizing $\mathbf{E}_2$ of Eq.~\ref{eq:error_s} using Gauss-Newton optimization. We use a stacked 8 dimension vector ${\mathbf{x} = [\mathbf{x}_1, \mathbf{x}_2]}$ to denote the object states at $1^{th}, 2^{th}$ frames, where $\mathbf{x}_i = [\mathbf{p}_i, \theta_i] \in \mathbb{R}^4$ (transpose is omitted for simplicity). During each iteration, we update the object states by
		\begin{equation}
		\label{eq:gn}
		\begin{array}{lr}
		\mathbf{x} \leftarrow \mathbf{x} + \Delta\mathbf{x},  \,\,\, \rm with \,\,\,\Delta \mathbf{x} = - \mathbf{H}^{-1}\mathbf{b}
		\end{array}
		\end{equation}
		where $\mathbf{H} \in \mathbb{R}^{8\times8}, \mathbf{b}\in \mathbb{R}^{8}$ are calculated by summarizing all costs and jacobians in Eq.~\ref{eq:error_s} respecting to the target states via standard Guass-Newton process. $\Delta\mathbf{x} \in \mathbb{R}^{8}$ is the state increment respecting to the current linearization point. After several iterations, we achieve an optimal estimation $\mathbf{x}$ obtained from a linear system:
		\begin{equation}
		\label{eq:gn2}
		\begin{array}{lr}
		\mathbf{x} = \widetilde{\mathbf{x}} + \Delta\mathbf{x},  \,\,\, \Leftrightarrow \,\,\,\mathbf{H}\mathbf{x} = \mathbf{H}\widetilde{\mathbf{x}} -\mathbf{b},
		\end{array}
		\end{equation}
		given by the last linearization point $\widetilde{\mathbf{x}}$ and the corresponding $\mathbf{H}, \mathbf{b}$. Eq.~\ref{eq:gn2} can be thought as a linear constraint for $\mathbf{x}$ that jointly considers two frames' stereo alignment, dense temporal correspondences, and individual pose constraints. Writing the linear constraints separately for two frames, we have
		\begin{equation}
		\label{eq:schur}
		\begin{array}{lr}
		\left[ {\begin{array}{cc}
			\mathbf{H}_{11} &\mathbf{H}_{12}\\
			\mathbf{H}_{21} &\mathbf{H}_{22}\\
			\end{array} } \right] \left[ {\begin{array}{c}
			\mathbf{x}_1\\
			\mathbf{x}_2\\
			\end{array} } \right] = \left[{\begin{array}{c}
			\mathbf{b}^\prime_{1} \\
			\mathbf{b}^\prime_{2} \\
			\end{array}}\right]  \,\,\, \rm with \,\,\,\mathbf{b}^\prime = \mathbf{H}\widetilde{\mathbf{x}} -\mathbf{b}
		\end{array}
		\end{equation}
		where $\mathbf{H}_{11}, \mathbf{H}_{22}$ contain the individual stereo and pose information for $1^{th}, 2^{th}$ frames, while $\mathbf{H}_{12}, \mathbf{H}_{21}$ symmetricly involve the temporal relations from dense pixel correspondences. Marginalizing the $\mathbf{x}_1$ from Eq.~\ref{eq:schur} using Schur complement will give us $\widetilde{\mathbf{H}} \mathbf{x}_2 = \widetilde{\mathbf{b}}^\prime$, derived by
		\begin{equation}
		\label{eq:schur2}
		\begin{array}{lr}
		\widetilde{\mathbf{H}} = \mathbf{H}_{22} - \mathbf{H}_{21}\mathbf{H}^{-1}_{11}\mathbf{H}_{12}; \,\,\,\,
		\widetilde{\mathbf{b}}^\prime = \mathbf{b}^\prime_{2} - \mathbf{H}_{21}\mathbf{H}^{-1}_{11}\mathbf{b}^\prime_{1}
		\end{array}
		\end{equation}
		As a result, we obtain an isolated linear constraint on the pose $\mathbf{x}_2$ of the $2^{th}$ frame while still taking both two frames' information into count. 
		
\begin{figure*}
		\begin{center}
			\includegraphics[width=2\columnwidth]{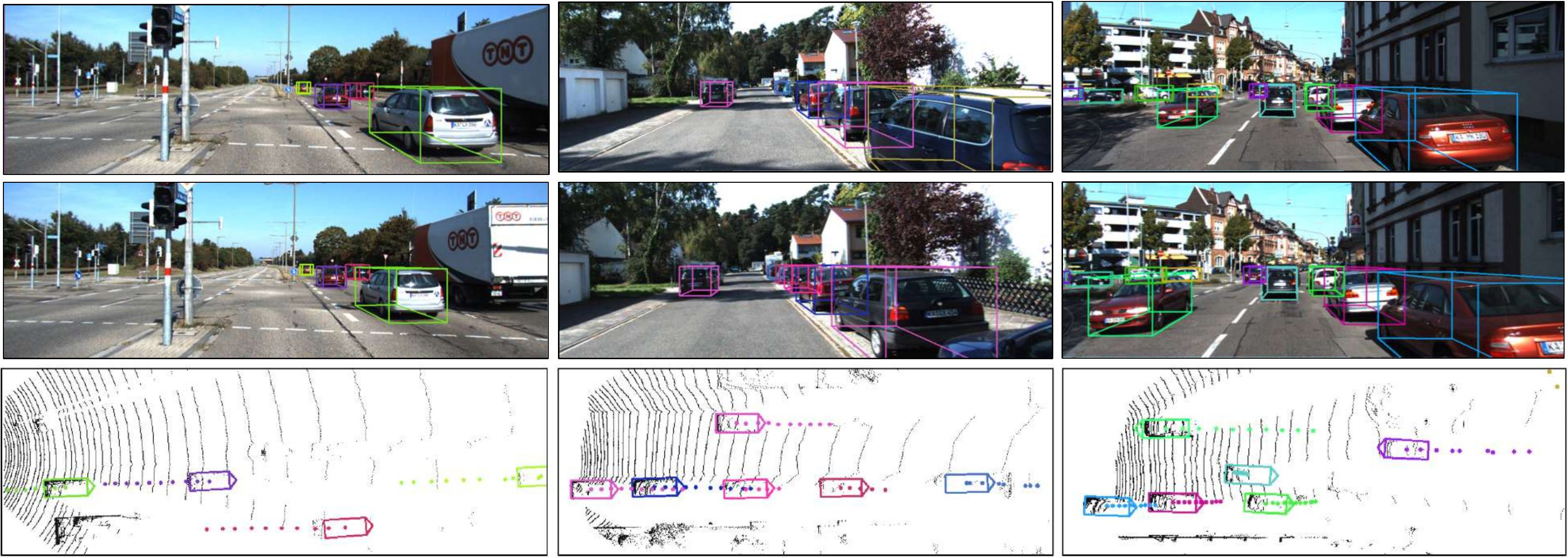}
		\end{center}
		\caption{\textbf{Qualitative results of our 3D object tracking.} We project the estimated 3D bounding box on two sequential images and bird's view map, where different color represents unique tracking id. Note that the color dots represent the \textit{relative} trajectory with respect to the corresponding ego-camera poses at each timestamp respectively.}
		\label{fig:quali}
	\end{figure*}

		When the object keep tracked in the $3^{th}$ frame, we can directly borrow the marginalized information as a prior to constrain the $2^{th}$ pose, meanwhile build the temporal constraints between $2^{th}$ and $3^{th}$ frame. Without loss of generality, for the $i^{th}$ frame we minimize 
		\begin{equation}
		\label{eq:final}
		\begin{array}{lr}
		\mathbf{E}_i = \mathbf{E}_{si} + \mathbf{E}_{ti} + \mathbf{E}_{pi} + \big\| \widetilde{\mathbf{H}} \mathbf{x}_{i-1} - \widetilde{\mathbf{b}}^\prime \big\|.
		\end{array}
		\end{equation}
		After $\mathbf{x}_i$ is solved, we analogously marginalize $\mathbf{x}_{i-1}$ as derivative in Eq.~\ref{eq:schur},\ref{eq:schur2} and extract the linear constraint for $\mathbf{x}_{i}$ that will be used for the next frame.
		In this way, we only need to evaluate the dense photometric error and temporal reprojection error for the current frame while still incorporate all history information. All previous stereo constraints will eventually contribute to the current estimation through step-by-step temporal relations. Note that our optimization solves a relative trajectory based on pure geometric relations, we thereby do not rely on the given ego-camera pose. Qualitative examples of our \textit{relative} trajectory estimation can be found in Fig.~\ref{fig:quali}.
	
	{\setlength{\parindent}{0cm}
		\subparagraph*{Alternative Way to Model Temporal Relations.} Besides finding dense pixel matching and minimizing the reprojection error in Eq.~\ref{eq:error2}, we also explore an alternative way to model the temporal relations by directly aligning the object local coordinates patch in adjacent frames, given by:
		\begin{equation}
		\label{eq:error_direct}
		\begin{array}{lr}
		\mathbf{E}_{ti} := \displaystyle\sum_{\mathbf{u}_i\in\mathcal{N}_t}w_{\mathbf{c}} \Big\| C_i(\mathbf{u}^p_{i}) - C_{i-1}(\mathbf{u}_{i-1}) \Big\|_h,
		\end{array}
		\end{equation}
		where the $C_i, C_{i-1}$ are the foreground local coordinates map in the current and previous frames respectively. 
Benefit from our learned local coordinates representation, we can evaluate a smooth gradient on $C_i, C_{i-1}$ map, and are robust to the exposure imbalance in different frames.  We compare and analysis these two ways in the experiment section (Table~\ref{table:3d},~\ref{table:tra} and Fig.~\ref{fig:error}).
	}
	
	\section{Implementation Details}
	{\setlength{\parindent}{0cm}
		\subparagraph*{Network Details.} We use ResNet-50 \cite{he2016deep} and FPN \cite{lin2017feature} as our network backbone. Three sets of anchor ratios \{0.5, 1, 2\} with four scales \{4, 8, 16, 32\} are used in the paired RPN stage. For each anchor, we regress 8-d outputs that correspond to the offsets for the box in the current and previous image respectively. 
		For the 2D box regression and 3D box estimation, we fed the concatenated RoI feature maps into two sequential fully-connected layers to produce 1024-d feature vectors. Similarly, we have 8 channels outputs for the paired 2D box regression and 6 channels output for the 3D box centroid and dimension regression. Following \cite{mousavian20173d}, we use two \textit{bin}s for angle classification and residual regression. For the dense prediction, we employ six stacked 256-d FCN (each layer is with 3 $\times$ 3 convolution and ReLU) on the dense RoI feature map, and predict 5-d dense output (1-d mask classification, 1-d local depth and 3-d local coordinates regression). The network inference time is $\sim$80 ms and the joint optimization takes $\sim$130 ms.
	}
	
	{\setlength{\parindent}{0cm}
		\subparagraph*{Training.} As MOTS \cite{Voigtlaender2019CVPR} provides dense instance segmentation labels for the KITTI tracking \cite{Geiger2012CVPR} sequences, we can directly use it for object mask supervision. The ground-truth for the local depth and local coordinates are calculated from the sparse LiDAR scanning and the labeled 3D bounding box. To leverage the network learn a better 3D object estimation, we firstly pretrain our model on the KITTI object dataset but excluded images appeared in the KITTI tracking sequences ($\sim$ 4000 images are left). Since the KITTI object dataset only provides single image with 3D object label, we apply two random scales with opposite direction (e.g., 0.95, 1.05) on the original image then crop or pad them into the original size, which can be roughly thought as equivalent scales in 3D object position. We thus get simulated adjacent images to initialize our tracking network. After that, we train our network on the tracking dataset to learn more actual association patterns. We expand the training set to 2$\times$ by flipping each image respecting the vertical axis, where the object angle and local coordinates are also mirrored respectively. 
		The total loss is defined as:
		\begin{equation}
		\label{eq:loss}
		\renewcommand*{\arraystretch}{1.5}
		\begin{array}{lr}
		{L} = \lambda_1L_{\rm rpn} + \lambda_2L_{\rm box} + \lambda_3L_{3d} + \lambda_4L_{\rm angle} + \lambda_5L_{\rm dense},
		\end{array}
		\end{equation}
		where $L_{\rm rpn}, L_{\rm box}, L_{\rm angle}, L_{\rm dense}$ contain both classification loss and regression loss, $\lambda_i$ denotes the individual uncertainty to balance the loss according to \cite{kendall2017multi}. 
		For each iteration, we feed one adjacent images pair into the network and sample 512 RoIs in RCNN stage.
		The network is trained using SGD optimizer with a momentum of 0.9 and
		a weight decay of 0.0005.  We train 10 epochs with 0.001 learning rate followed by 2 epochs with 0.0001 learning rate.
	}
	
	\begin{table*}
		\begin{center}
			\renewcommand{\arraystretch}{1.3}
			\resizebox{0.93\textwidth}{!}{%
				\begin{tabular}{l|c|ccccccc|ccccccc}
					\, & \, &
					\multicolumn{7}{c|}{3D IoU = 0.25} & \multicolumn{7}{c}{{3D IoU = 0.5}} \\
					\cline{3-16}
					Method & Sensor &  MOTA $\uparrow$ & MOTP $\uparrow$ & F1 $\uparrow$ & MT $\uparrow$ & ML $\downarrow$ &  \# FP $\downarrow$ &  \# FN $\downarrow$ & MOTA $\uparrow$ & MOTP $\uparrow$  & F1 $\uparrow$ & MT $\uparrow$ & ML $\downarrow$ & \# FP $\downarrow$ & \# FN $\downarrow$  \\
					\Xhline{1pt}
					Joint-Tracking \cite{hu2018joint} & Mono & -15.55 \footnotemark[1] & 47.91 & 42.14 & 9.33 & 33.33 & 3855 & 4868 & -55.57 & 63.76 & 18.90 & 0.67 & 68.00 & 5378 & 6366 \\
					Semantic-Tracking \cite{li2018stereo} & Stereo & 3.31 & 51.72 & 47.32 & 11.33 & 40.67 & 2662 & 4620 & -34.14 & 65.39 & 24.72 & 2.00 & 62.67 & 4070 & 6054 \\
					\cline{1-16} 
					Ours (Coord) & Stereo & 56.14 & 62.20 & 77.53 & 42.67 & 14.00 & 820 & 2464 & 28.56 & 69.34 & 61.67 & 22.67 & 24.00 & 1730 & 3651 \\
					Ours (Repro) & Stereo & \textbf{56.70} & \textbf{62.33} & \textbf{77.85} & \textbf{44.00} & \textbf{12.00} & \textbf{794} & \textbf{2443} & \textbf{29.39} & \textbf{69.39} & \textbf{62.13} & \textbf{24.00} & \textbf{23.33} & \textbf{1697} & \textbf{3618} \\
				\end{tabular}
			}		
		\end{center}
		\caption{\textbf{3D bounding box tracking results on the KITTI tracking \textit{val} set}, where 3D box IoU are used for True Positive (TP) assignments.}
		\label{table:3d}
	\end{table*}
	
	\begin{table*}
		\begin{center}
			\renewcommand{\arraystretch}{1.3}
			\resizebox{0.99\textwidth}{!}{%
				\begin{tabular}{l|c|ccccc|ccccc|ccccc}
					\, & \, &
					\multicolumn{5}{c|}{Distance = 3m} & \multicolumn{5}{c|}{{Distance = 2m}} & \multicolumn{5}{c}{{Distance = 1m}} \\
					\cline{3-17}
					Method & Sensor &  MOTA $\uparrow$ & MOTP $\downarrow$ & F1 $\uparrow$ & MT $\uparrow$ & ML $\downarrow$ &  MOTA $\uparrow$ & MOTP $\downarrow$  & F1 $\uparrow$ & MT $\uparrow$ & ML $\downarrow$ & MOTA $\uparrow$ & MOTP $\downarrow$ & F1 $\uparrow$ & MT $\uparrow$ & ML $\downarrow$  \\
					\Xhline{1pt}
					3D-CNN/PMBM \cite{scheidegger2018mono}  & Mono & 47.20 & 1.11 m & 73.86 & 48.65 & 11.35  & - & - & - & - & - & - & - & - \, \\
					Joint-Tracking \cite{hu2018joint} & Mono & 47.22 & 1.13 m & 75.63 & 40.00 & 7.33 & 27.16 & 0.88 m & 65.20 & 28.00 & 12.67 & -14.58 & 0.53 m & 42.52 & 10.67 & 33.33 \\
					Semantic-Tracking \cite{li2018stereo} & Stereo & 51.19 & 1.00 m & 74.82 & 39.33 & 12.00  & 34.84 & 0.76 m & 65.54 & 28.67 & 20.00 & 4.03 & 0.49 m & 47.76 & 14.00 & 38.67 \\
					\cline{1-17}
					Ours (Coord) & Stereo & 74.75 & 0.49 m & 87.66 & 64.67 & 7.33 & 71.12 & 0.44 m & 85.69 & 58.67 & 8.67 & 56.11 & 0.32 m & 77.57 & 43.33 & 13.33 \\
					Ours (Repro) & Stereo & \textbf{74.92} & \textbf{0.49 m} & \textbf{87.77} & \textbf{65.33} & \textbf{7.33} & \textbf{71.40} & \textbf{0.44 m} & \textbf{85.85} & \textbf{60.67} & \textbf{8.00} & \textbf{56.74} & \textbf{0.32 m} & \textbf{77.94} & \textbf{47.33} & \textbf{12.00} \\
				\end{tabular}
			}		
		\end{center}
		\caption{\textbf{3D trajectory tracking results on the the KITTI tracking \textit{val} set.} We assign the True Positive trajectories according to the 3D Euclidean distance between object centroids with different threshholds. Note that the tracking precision (MOTP) is defined based on the distance error, i.e., the lower the better.}
		\label{table:tra}
	\vspace{-1mm}
	\end{table*}
	
	\section{Experiments}
	\label{sec:exp}
	We evaluate our method on the KITTI tracking dataset \cite{Geiger2012CVPR} using the standard CLEAR \cite{Bernardin2008} metric for multiple objects tracking (MOT) evaluation. As this paper focuses on the 3D object tracking, we define the similarity function between the estimated trackers and ground truth objects in the 3D space. Specifically, we use the overlap between two 3D object cuboids with 0.25 and 0.5 IoU thresholds to evaluate the 3D bounding box tracking performance, and use the Euclidean distance between 3D object centroids to evaluate the 3D trajectory tracking performance (3, 2, 1 meters thresholds are used respectively). 
	The overall tracking performance is evaluated by the MOTA (multiple objects tracking accuracy), MOTP (multiple objects tracking precision), F1 score (calculated from the precision and recall), MT (most tracked percent), ML (most lost percent), and FP (false positives), FN (false negatives), etc. Since the official KITTI tracking server does not support 3D tracking evaluation, we follow \cite{Voigtlaender2019CVPR} to split the whole train data into \textit{training} and \textit{val} set, and conduct extensive comparisons and ablation analysis on the val set for the car category. We also report 3D pedestrian tracking results and extend the evaluation to KITTI raw \cite{Geiger2013IJRR} and Argoverse tracking \cite{chang2019argoverse} dataset.
	\begin{table}
		\begin{center}
			\renewcommand{\arraystretch}{1.4}
			\resizebox{0.41\textwidth}{!}{%
				\begin{tabular}{l|ccccc}
					Method & MOTA $\uparrow$ &MOTP $\uparrow$ & F1 $\uparrow$ & MT $\uparrow$ & ML $\downarrow$ \\
					\Xhline{1pt}
					Stereo R-CNN \cite{li2019stereo} & 23.59 & 69.98 & 56.29 & 18.00 & 28.00   \\
					Pseudo-LiDAR \cite{wang2019pseudo} & 25.88 & \textbf{71.10} & 58.14 & 20.00 & 25.33  \\
					\cline{1-6}
					Ours (Coord) & 28.56 & 69.34 & 62.13 & 24.00 & 23.33 \\
					Ours (Repro) & \textbf{29.39} & 69.39 & \textbf{62.13} & \textbf{24.00} & \textbf{23.33}  \\
				\end{tabular}
			}		
		\end{center}
		\caption{\textbf{Comparing with 3D detectors + KF tracker \cite{weng2019baseline}}. Note that MOTP \cite{bernardin2008evaluating} is defined on TPs (3D IoU $>$ 0.5) only, which is independent of the overall tracking a consistent trajectory ability. }
		\label{table:det}
		\vspace{-2mm}
	\end{table}
	\vspace{-4mm}
	{\setlength{\parindent}{0cm}
		\subparagraph*{3D Object Tracking Evaluation.} We compare our 3D bounding box and 3D trajectory tracking performance with recent image-based 3D object tracking approaches in Table.\ref{table:3d},~\ref{table:tra} respectively, where 3D-CNN/PMBM \cite{scheidegger2018mono} and Joint-Tracking \cite{hu2018joint} use monocular image for object detection and use PMBM filter or LSTM to generate continuous 3D tracking. Semantic-Tracking \cite{li2018stereo} uses stereo images to achieve a better 3D localization accuracy. As the code for PMBM \cite{scheidegger2018mono} is not available, we directly list the 3D trajectory tracking results in its original paper for reference. We finetune \cite{hu2018joint} on the same tracking split \cite{Voigtlaender2019CVPR} based on its released pre-trained weight on a large scale GTA dataset. For Semantic-Tracking \cite{li2018stereo}, we replace the 2D IoU-based association and the fixed size prior in its original implementation to our learned association and dimension for a fair comparison. As detailed in Table.~\ref{table:3d},~\ref{table:tra}, our method significantly outperforms all image-based 3D object tracking methods for both 3D bounding box and 3D trajectory tracking evaluation. Note that 3D MOTA can be negative\footnotemark[1] as it assigns TPs using 3D IoU or 3D distance, which poses a high strict requirement for image-based approaches.
		Although \cite{li2018stereo} employs the stereo sensor and considers the motion consistency as well, however, it solves the object relative motion in a sliding window and aligns the object box to the sparse point cloud recovered by the discrete stereo feature matching in separate stages, which is in essence differ from our joint spatial-temporal optimization approach. Both the sparse stereo matching and loosely coupled spatial-temporal information limit its 3D tracking performance.
	}
	\footnotetext[1]{$\rm MOTA = (1 - \dfrac{\sum (\rm FN + FP + IDS)}{\sum \rm GT} )\times100$, i.e. $\in (-\infty, 100)$}
	
	We also note that modeling temporal relations by local coordinates error in Eq.~\ref{eq:error_direct} (denoted as Coord) slightly underperforms the reprojection error in Eq.~\ref{eq:error2} (denoted as Repro). As minimizing the local coordinates error tries to align the whole object patch, however, the visible areas are not identical even for adjacent frames due to slight viewpoint changing and truncation. An error pattern to reveal the phenomenon can be found in Fig.~\ref{fig:error}, where we can observe a large error in the rear wheel region because the optimizer tries to align the truncated patch to the complete patch in the last frame. Minimizing reprojection error avoids this issue easily by setting a distance threshold for local coordinates matching. If not specified, we report our (repro) results in the following experiments by default.
	
	{\setlength{\parindent}{0cm}
		\subparagraph*{Comparison with 3D Detection Methods.}  To further demonstrate our tracking performance, we extend the comparison to state-of-the-art stereo 3D detection methods Stereo RCNN \cite{li2019stereo} and Pseudo LiDAR \cite{wang2019pseudo}. We train these two detectors on our KITTI object split which does not contain images in KITTI tracking sequences, and run the inference on the KITTI tracking $val$ set. We use the recent proposed KF-based tracker \cite{Weng2019_3dmot} to associate the discrete detections and produce sequential 3D object trajectories. As Table.~\ref{table:det} shows, although the detection-based method \cite{wang2019pseudo} shows a good precision (MOTP) for True Positives, a KF tracker cannot guarantee the optimal trajectory from only detection data as most of the original information is lost. We outperform them in the overall tracking performance (MOTA, MT, etc), which evidences again the advantage of our joint spatial-temporal optimization approach.
	}
	{\setlength{\parindent}{0cm}
		\subparagraph*{Benefits of Spatial \& Temporal Information.} This experiment shows how the spatial and temporal information helps our 3D object tracking. As listed in Table.~\ref{tab:aba}, we use the Regress to denote the 3D tracking result using the monocular regressed 3D box only, which shows inadequate 3D tracking performance. While modeling spatial constraints (stereo alignment) significantly improves the 3D localization ability due to introducing accurate depth-sensing ability. Further, adding temporal information by considering geometric relations and motion consistency improves 3D tracking robustness again. The tracking accuracy (MOTA), tracking precision (MOTP) and tracking robustness (MT, ML) are all improved by remarkable margins.
	}

	\begin{table}
		\begin{center}
		\renewcommand{\arraystretch}{1.4}
		\resizebox{0.4\textwidth}{!}{%
		\begin{tabular}{l|ccccc}
			Method & MOTA $\uparrow$ &MOTP $\uparrow$ & F1 $\uparrow$ & MT $\uparrow$ & ML $\downarrow$ \\
			\Xhline{1pt}
			Mono Regress & -35.26 & 66.96 & 22.28 & 0.00 & 64.00   \\
			+ Spatial & 26.86 & 69.24 & 60.62 & 18.00 & 24.00  \\
			+ Temporal & \textbf{29.39} & \textbf{69.39} & \textbf{62.13} & \textbf{24.00} & \textbf{23.33}  \\
		\end{tabular}
	}		
\end{center}
		\caption{\textbf{Comparing effects of adding different information}.}
		\label{tab:aba}
		\vspace{-2mm}
	\end{table}
	{\setlength{\parindent}{0cm}
		\subparagraph*{More Quantitative Experiments.} Since our method predicts object shape and is based on pure geometry, we can seamlessly use it for 3D pedestrian tracking. The quantitative results on the KITTI tracking set and an example can be found in Table.~\ref{tab:ped} and Fig.~\ref{fig:ped} respectively. Besides the evaluation on the KITTI tracking dataset, we also report our 3D tracking results on the KITTI raw sequence \cite{Geiger2013IJRR} and Argoverse Tracking \cite{chang2019argoverse} dataset for future benchmarking.  As reported in Table.~\ref{tab:more}, we evaluate on totall 24 KITTI raw sequences that are excluded from the tracking dataset. Note that here we train the network on the whole KITTI tracking dataset without pretraining on the object dataset as the KITTI object images are distributed in most of the raw sequences.  The Argoverse dataset provides stereo images with 5 fps and labeled 3D object trackers on 10 fps LiDAR scans. Since the official server only evaluates the 10 fps 3D object tracking on the LiDAR timestamps, we thereby report our results on the 24 stereo validation sequences by assigning the ground truths of the LiDAR frame with the nearest timestamp. As detailed in Table.~\ref{tab:more}, we note that our image-based method works reasonably in short range while unavoidably suffers from performance decent for long-range objects. This is due to a combined reason for low fps stereo images, reciprocal relations between disparity and depth, and non-trivial projection error for extremely faraway objects, etc. 
	}
	
	\section{Conclusion}
	In this paper, we propose a joint spatial-temporal optimization approach for stereo 3D object tracking. Our method models the relations between the invisible object centroid and the local object geometric cues into a joint spatial photometric and temporal reprojection error function. By minimizing the joint error with a per-frame marginalized prior, we estimate an optimal object trajectory that satisfies both the instant stereo constraints and accumulated history evidence.
	Our approach significantly outperforms previous image-based 3D tracking methods on the KITTI tracking dataset. Extensive experiments on multiple categories and larger datasets (KITTI raw and Argoverse Tracking) are also reported for future benchmarking.

	\begin{figure}
		\begin{center}
			\includegraphics[width=1.0\columnwidth]{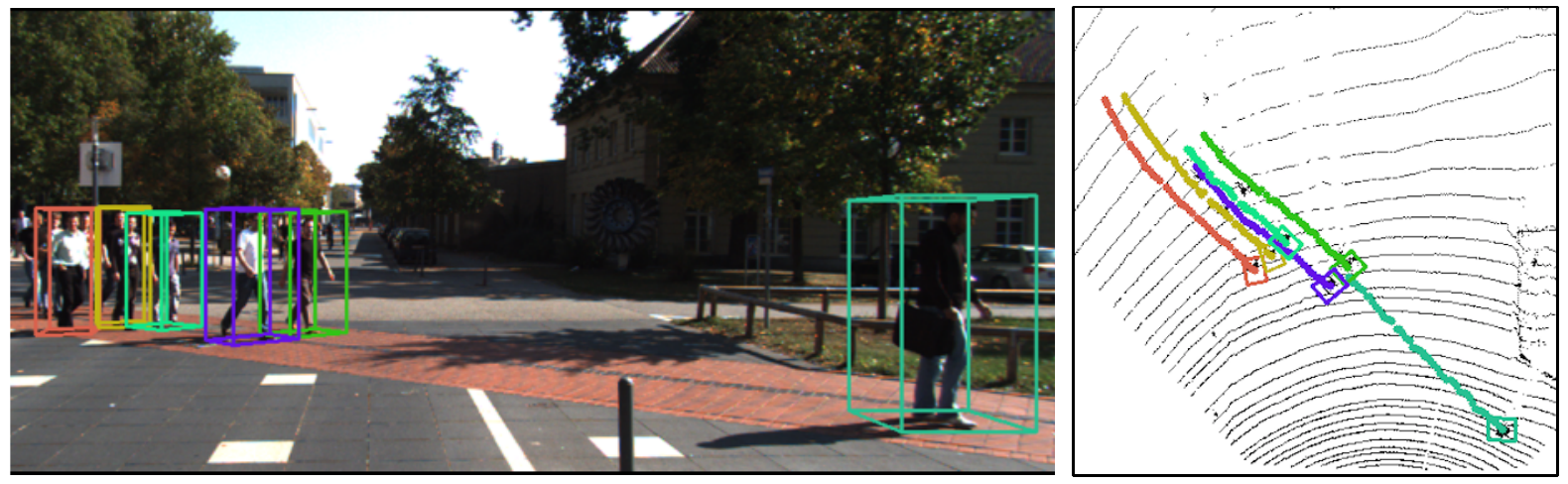}
		\end{center}
		\caption{\textbf{Qualitative example for 3D pedestrian tracking}.}
		\label{fig:ped}
	\end{figure}
	
	\begin{table}
		\begin{center}
			\renewcommand{\arraystretch}{1.3}
			\resizebox{0.47\textwidth}{!}{%
				\begin{tabular}{l|ccccccc}
					Threshhold & MOTA & MOTP & F1 & MT & ML & \# FP\, & \# FN \,\\
					\Xhline{1pt}
					Distance = 1 m  & 33.79 & 0.26 m & 67.78 & 44.12 & 13.24 & 1014 & 1082\\
					3D IoU = 0.25 & 16.73 & 48.02 & 58.60 & 27.94 & 22.06 & 1276 & 1392\\
			\end{tabular}}
		\end{center}
		\caption{\textbf{3D pedestrian tracking results on KITTI tracking \textit{val} set}. Note that we require the true positive trajectory has $< 1$ m distance error since pedestrians are more crowded than vehicles.}
		\label{tab:ped}
	\end{table}
	
	\begin{table}
		\begin{center}
			\renewcommand{\arraystretch}{1.3}
			\resizebox{0.46\textwidth}{!}{%
				\begin{tabular}{l|l|ccccc}
					Dataset & Threshhold & MOTA & MOTP & F1 & MT & ML \,\\
					\Xhline{1pt}
					\multirow{2}{*}{KITTI Raw} & Distance = 2 m  & 63.02 & 0.47 m & 84.81 & 50.32 & 14.95 \\
					
					\, & 3D IoU = 0.25 & 46.29 & 59.88 & 77.07 & 37.89 & 21.05\\
					\hline
					\multirow{3}{*}{\shortstack{Argoverse\\Tracking}} & Range 100 m & 4.10 & 0.93 m & 46.30 & 16.09 & 40.66 \\
					\, & Range 50 m & 25.71 & 0.87 m & 63.00 & 30.72 & 21.02 \\
					\, & Range 30 m & 43.81 & 0.68 m & 76.24 & 72.92 & 7.22\\
			\end{tabular}}
		\end{center}
		\caption{\textbf{Evaluation on KITTI raw sequences and Argoverse datasets}, where we seperate the evaluation range and use 2.25 m 3D centroid distance as the threshold for true positive assignment following the Argoverse official setting.}
		\label{tab:more}
		\vspace{-2mm}
	\end{table}

	
{\setlength{\parindent}{0cm}		
	\subparagraph*{Acknowledgment.} This work was supported by the HKUST Postgraduate Studentship and the Hong Kong Research Grants Council Early Career Scheme under project 26201616.		
}	
	
	{\small
		\bibliographystyle{ieee_fullname}
		\bibliography{egbib}
	}

\begin{figure*}[t]
	\begin{center}
		\includegraphics[width=2.0\columnwidth]{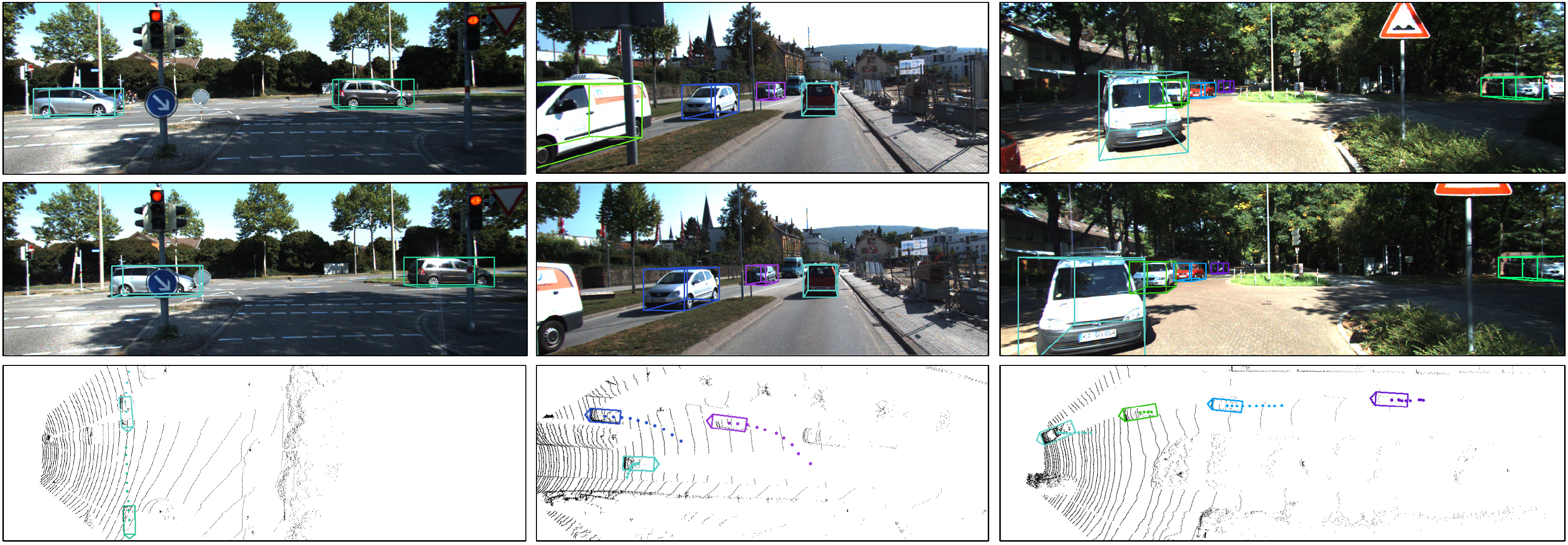}
	\end{center}
	\caption{\textbf{More qualitative results on the KITTI dataset.} Note that the relative trajectories of 3D objects respecting to the ego-camera are visualized on the bird's eye view image.}
	\label{fig:kitti}
\end{figure*}

\begin{figure*}
	\begin{center}
		\includegraphics[width=2.0\columnwidth]{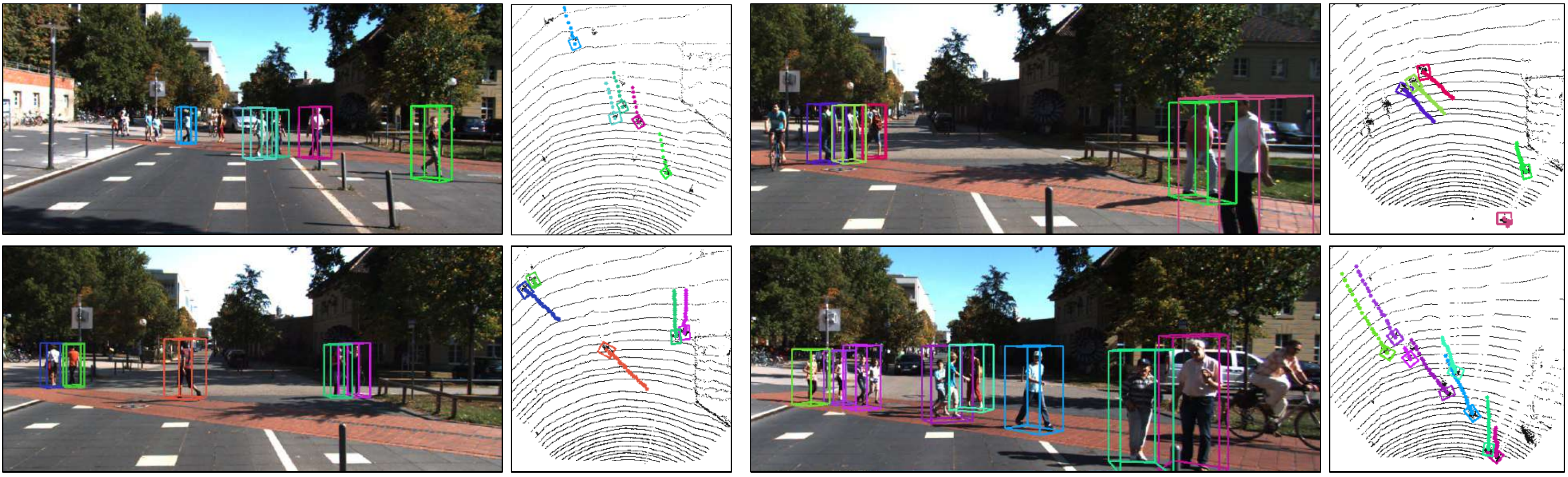}
	\end{center}
	\caption{\textbf{More qualitative results of the 3D pedestrian tracking on the KITTI dataset.} Each bird's eye view image corresponds to its left RGB image.}
	\label{fig:ped}
\end{figure*}	

\section*{Appendix}

{\setlength{\parindent}{0cm}
	\subparagraph*{2D Tracking Evaluation and Analysis.} 
Although this paper mainly focuses on the 3D object tracking, we submit our results to the KITTI 2D tracking test server to provide readers a complete reference to our 3D tracking system. As we introduced in Sect 3.1, to make our framework simple and efficient, we extend the joint stereo proposal strategy in \cite{li2019stereo} to the sequential images, which enables us simultaneously detect and associate 2D objects without additional pair-wise similarity computation. Though simple, our 2D tracker demonstrates good tracking performance compared to recent state-of-the-art 2D tracking methods as shown in Table.~\ref{table:2d}.  
}
	
An interesting phenomenon is that our 2D tracker produces lowest False Positives (FP) and higher False Negatives (FN) compared to \cite{zhang2019robust, hu2018joint, sharma2018beyond}, which can be explained by the characteristic of the paired proposal. Since only the anchor which overlaps with the union area of the sequential 2D object box will be treated as a foreground proposal, i.e., the ``alarm threshold" for positive samples is increased, which significantly reduces the false positive rate. Similarly, due to the variant location (nearby large objects, fast motion, etc) of the object on adjacent images, a set of predefined anchors may miss covering some distantly located pairs, which can be potentially overcome with the help of recent anchor-free 2D detection approaches \cite{tian2019fcos}.  Although slightly underperform  \cite{hu2018joint} in 2D tracking, we show significant better 3D tracking performance benefit from our joint spatial-temporal optimization. Employing additional object similarity calculation or exploring anchor-free based paired proposal may further boost our association performance, while outside the main scope of this work.

\begin{figure*}
	\begin{center}
		\includegraphics[width=2.05\columnwidth]{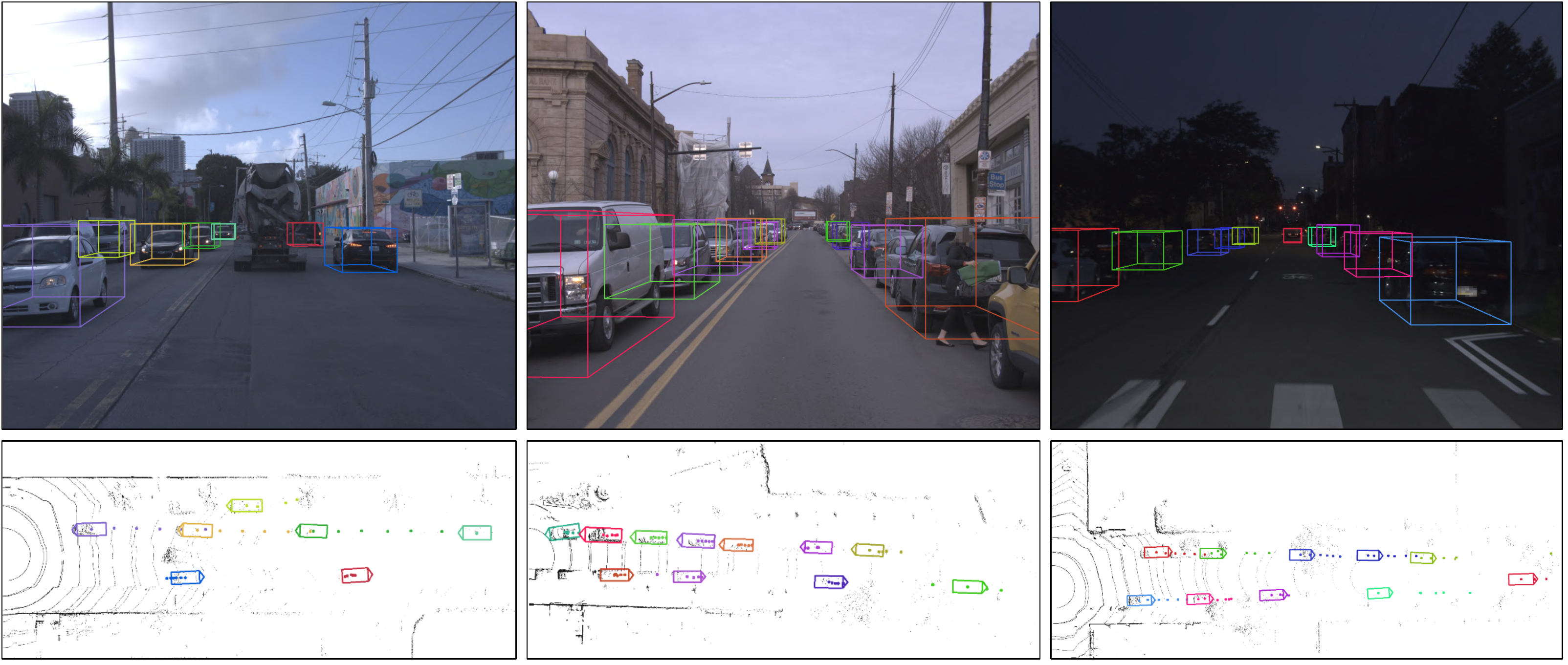}
	\end{center}
	\caption{\textbf{qualitative results on the Argoverse Tracking dataset,} where the stereo images are recored in 5 fps with small FOV cameras.}
	\label{fig:argo}
\end{figure*}

\begin{table}
	\begin{center}
		\renewcommand{\arraystretch}{1.4}
		\resizebox{0.48\textwidth}{!}{%
			\begin{tabular}{l|cccccc}
				Method & MOTA $\uparrow$ &MOTP $\uparrow$ & MT $\uparrow$ & ML $\downarrow$ & \# FP $\downarrow$ & \# FN $\downarrow$ \\
				\Xhline{1pt}
				mmMOT \cite{zhang2019robust} & \textbf{84.77} &	85.21  &73.23  & \textbf{2.77} & 711 & 4243   \\
				Joint-Tracking \cite{hu2018joint} & 84.52 &	85.64  & \textbf{73.38} & 2.77  & 705 & \textbf{4242} \\
				MOTBeyondPixels \cite{sharma2018beyond} & 84.24  & \textbf{85.73} & 73.23  & 2.77 & 705 & 4247 \\
				JCSTD \cite{tian2019online} & 80.57 &	81.81 &	56.77  & 7.38 & 405	& 6217 \\
				3D-CNN/PMBM \cite{scheidegger2018mono} & 80.39 &	81.26 &	62.77 &	6.15 & 1007	& 5616 \\
				FAMNet \cite{Chu_2019_ICCV} & 77.08  &	78.79  & 51.38  & 8.92  &  760 &	6998	\\
				\cline{1-7}
				Ours (ST-3D) & 82.64 &	83.83  &61.69  & 7.23  & \textbf{234} & 5366 \\
			\end{tabular}
		}		
	\end{center}
\caption{\textbf{2D tracking results on the KITTI test set.} We mainly list recently published methods (whether they have the 3D tracking ability or not) for reference. }
\label{table:2d}
\vspace{-3mm}
\end{table}
\vspace{-3mm}
{\setlength{\parindent}{0cm}
	\subparagraph*{More Qualitative Examples.} 
	We visualize more qualitative examples on KITTI 3D car and 3D pedestrian tracking and Argoverse Tracking in Fig.~\ref{fig:kitti}, \ref{fig:ped}, \ref{fig:argo} respectively, where the relative trajectories on the bird's eye view are also showcased.
}

\end{document}